\documentclass[10pt,twocolumn,letterpaper]{article}

\usepackage{cvpr}
\usepackage{times}
\usepackage{epsfig}
\usepackage{graphicx}
\usepackage{amsmath}
\usepackage{amssymb}

\usepackage{caption}
\usepackage{authblk}
\usepackage{booktabs,tabularx}	
\usepackage{lipsum}
\usepackage{url}

\usepackage[pagebackref=true,breaklinks=true,letterpaper=true,colorlinks,bookmarks=false]{hyperref}

\cvprfinalcopy 

\def\cvprPaperID{} 

\ifcvprfinal\fi

\begin{document}

\title{Text Guided Person Image Synthesis}
\author{Xingran Zhou\textsuperscript{1} \hspace{.04in} Siyu Huang\textsuperscript{1}\thanks{} \hspace{.04in} Bin Li\textsuperscript{1} \hspace{.04in} 
\vspace{.05in}
Yingming Li\textsuperscript{1} \hspace{.04in} Jiachen Li\textsuperscript{2} \hspace{.04in} Zhongfei Zhang\textsuperscript{1}\\
\textsuperscript{1}\thinspace Zhejiang University \hspace{.1in} \vspace{.03in}\textsuperscript{2}\thinspace Nanjing University
\vspace{-.1in}
{\tt\small {\{xingranzh,~siyuhuang,~bin\_li,~yingming,~zhongfei\}@zju.edu.cn}, jiachen\_li\_nju@163.com}
}

\twocolumn[{%
\renewcommand\twocolumn[1][]{#1}%
\vspace{-2em}
\maketitle
\thispagestyle{empty}
\vspace{-2em}
\begin{center}
\centering
\includegraphics[width=\linewidth]{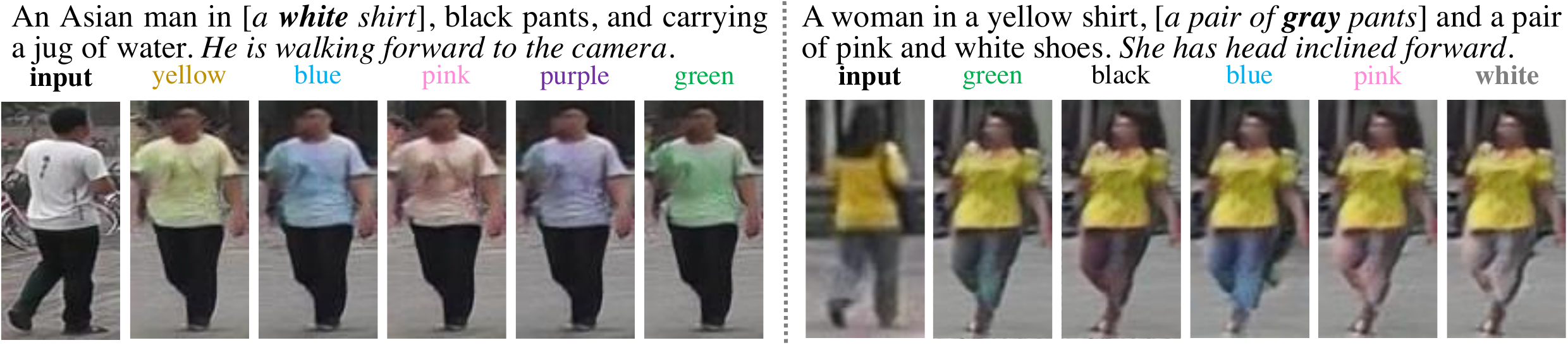}
\vspace{-1em}
\captionof{figure}{\textbf{Samples of text guided person image synthesis.} Given the reference images and the natural language descriptions, our algorithm correspondingly generates pose and attribute transferred person images. As shown in the left, our algorithm transfers the person pose based on \textit{`He is walking forward to the camera'}, and also synthesizes shirts of various different colors. Similarly for the right example.
}
\label{fig:abstract_fig}
\end{center}%
}]
\newcommand\blfootnote[1]{%
\begingroup 
\renewcommand\thefootnote{}\footnote{#1}%
\addtocounter{footnote}{-1}%
\endgroup 
}

\begin{abstract}
This paper presents a novel method to manipulate the visual appearance (pose and attribute) of a person image according to natural language descriptions. Our method can be boiled down to two stages: 1) text guided pose generation and 2) visual appearance transferred image synthesis. In the first stage, our method infers a reasonable target human pose based on the text. In the second stage, our method synthesizes a realistic and appearance transferred person image according to the text in conjunction with the target pose. Our method extracts sufficient information from the text and establishes a mapping between the image space and the language space, making generating and editing images corresponding to the description possible. We conduct extensive experiments to reveal the effectiveness of our method, as well as using the VQA Perceptual Score as a metric for evaluating the method. It shows for the first time that we can automatically edit the person image from the natural language descriptions.
\end{abstract}
\blfootnote{* Corresponding author}

\vspace{-5pt}
\section{Introduction}
\vspace{-5pt}
Person images are produced in any time today due to the popularization of visual capturing devices such as mobile phones, wearable cameras, and surveillance systems. The demand for a user-friendly algorithm to manipulate person images is growing rapidly. In practice, people usually represent the concept about a person's appearance and status in a very flexible form, \ie, the natural languages. It is our understanding that the guidance through text description for generating and editing images is a friendly and convenient way for person image synthesis.

In this paper, we propose a new task of \emph{editing a person image according to natural language descriptions}. Two examples of this task are shown in Fig.~\ref{fig:abstract_fig}. Given an image of a person, the goal is to transfer the visual appearance of the person under the guidance of text description, while keeping the invariance of person identity. Specifically, the pose, attributes (\eg, cloth color), and the other properties of the identity are simultaneously edited to satisfy the description.

Generative Adversarial Networks (GANs) have offered a solution to conditioned realistic image generation. The text-to-image approaches~\cite{Tao18attngan, reed2016generative, zhang2018photographic, zhang2017stackgan, zhang2017stackgan++} synthesize images with given texts without the reference images, where the semantic features extracted from texts are converted to the visual representations to constitute the generated images. However, most of these approaches only succeed in flower or bird image generation. In regard to person editing, the pose guided generation methods~\cite{ma2017pose, ma2018disentangled, esser2018variational, balakrishnan2018synthesizing} transfer the person pose by taking the target pose as input to instruct the generation process, while the editing of person image under the guidance of natural language descriptions has rarely been studied in the existing literature.

\begin{figure}[t]
\begin{center}
\includegraphics[width=\linewidth]{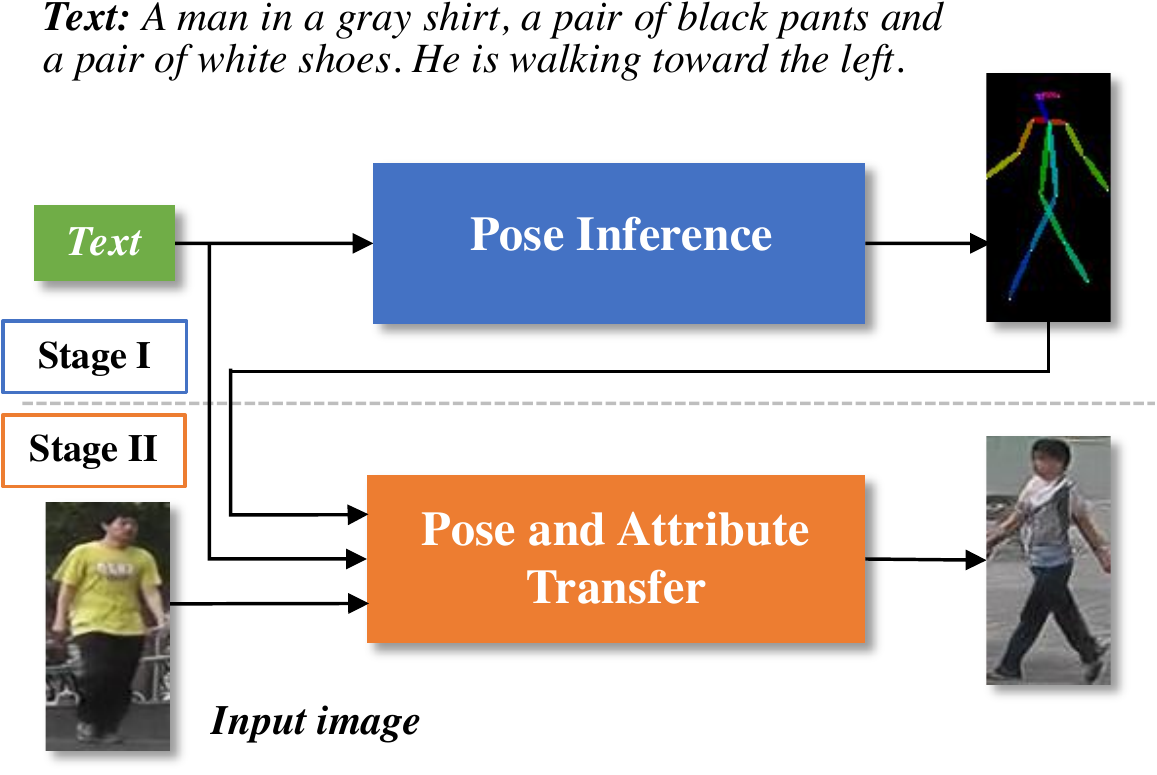}
\end{center}
\vspace{-8pt}
\caption{\textbf{Simple illustration of our approach.} In Stage-I, we infer a reasonable human pose from the natural language description. In Stage-II, our method takes the predicted pose, the reference image, and the text as input to synthesize a pose and attribute transferred person image by keeping the person identity.
}
\vspace{-10pt}
\label{fig:intro_fig}
\end{figure}

Motivated by these considerations, we propose a novel text guided person image synthesis framework, which is able to semantically edit the pose and the attributes of the person in consistence with the text description while retaining the identity of the person. As shown in Fig.~\ref{fig:intro_fig}, our method is comprised of two stage successively. The two stages are both built upon the adversarial learning conditioned on the text description. Specifically, Stage-I is a newly proposed pose inference network, in which a reasonable target pose is inferred from the text description as an intermediate product to lay the foundation for the subsequent pose transferring. A set of basic poses is drawn from the training dataset. The pose inference network first selects a basic pose with respect to the exact direction, and then refines every joint in details to conform to the text description. By the pose inference network, the target pose is guaranteed to model the shape and the layout of unique body pose structure of a person.

Stage-II takes the predicted pose, the reference image, and the text description as input to yield a realistic-look pedestrian image by manipulating both the pose and the appearance of the reference image. A multi-modal learning loss involved an attention mechanism is proposed to establish the link among different words in the text and the sub-regions in the image. Moreover, a novel attention upsampling module is developed in this stage for better combining the pose feature and the semantic embedding. Compared with the previous image editing methods, our model is able to simultaneously manipulate multiple person attributes, enabling a more interactive and flexible approach to person image synthesis.

The contribution of this paper is listed as follows. 1) We propose a new task of manipulating person images based on natural language descriptions, towards the goal of user-friendly image editing. 2) For the first time, we propose a GAN-based pose inference network to generate the human pose according to the text description, to the best of our knowledge. 3) We present a novel two-stage framework for text guided person image synthesis, in which the modules of attention upsampling and multi-modal loss are introduced to establish semantic relationships among images, poses, and natural language descriptions. 4) We propose the VQA Perceptual Score to evaluate the correctness of attribute changes corresponding to specific body parts.

\vspace{-5pt}
\section{Related works}
\vspace{-5pt}
\noindent
\textbf{Deep generative models. }
In recent years, deep generative models including Generative Adversarial Networks (GANs)~\cite{goodfellow2014generative}, Variational Auto-encoders (VAEs)~\cite{kingma2013auto}, and Autoregressive (AR) models~\cite{van2016conditional} have attracted wide interests in the literature. The advances of generative models also drive further studies on generating and translating images, including the image to image translation~\cite{isola2017image, zhu2017unpaired, choi2017stargan}, super-resolution~\cite{ledig2017photo, odena2016conditional, karras2017progressive}, and the style transfer~\cite{zhang2016colorful, zhu2017toward, karacan2016learning}. These techniques are of great importance for the computer vision research and with a plenty of applications.

\noindent
\textbf{Person image generation. }
Recent work has achieved impressive results in generating person images in the expected poses. For instance, Ma \etal~\cite{ma2017pose} propose Pose Guided Person Image Generation ($\text{PG}^\text{2}$) which initially generates a coarse image and then refines the blurry result in an adversarial way. Balakrishnan \etal~\cite{balakrishnan2018synthesizing} present a modular generative network which separates a scene into various layers and moves the body parts to the desired pose. Ma \etal~\cite{ma2018disentangled} use a disentangled representation of the image factors (foreground, background, and pose) to composite a novel person image. Esser \etal~\cite{esser2018variational} present a conditional U-Net shape-guided image generator based on VAE for person image generation and transfer. It is desirable to edit and manipulate person images according to natural language descriptions.

\begin{figure*}[!ht]
\begin{center}
\includegraphics[width=\linewidth]{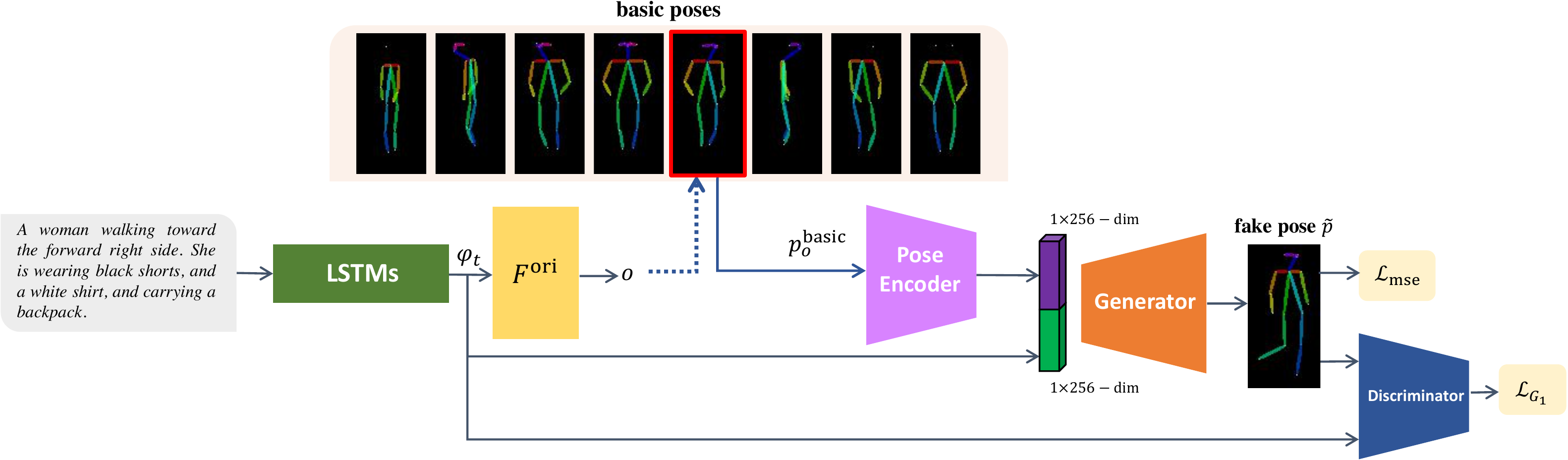}
\end{center}
\vspace{-13pt}
\caption{\textbf{Stage-I: Text guided pose generator.} We conclude the prior about poses in the training dataset as a series of basic poses. We first predict the orientation of the pose from the text by the orientation selection net $F^{o}$. Then, we train a single generator $G1$ that learns to manipulate every joint in the pose for fine-grained details.
}
\label{fig:model_1}
\end{figure*}

\noindent
\textbf{Text conditioned generation. }
Reed \etal~\cite{reed2016generative} first propose an end-to-end architecture based on conditional GANs framework, which generates realistic $64 \times 64$ images for birds and flowers from natural language descriptions. Their follow-up work~\cite{reed2016learning} is able to generate $128 \times 128$ images by incorporating additional annotations of object parts. The StackGAN~\cite{zhang2017stackgan, zhang2017stackgan++} is proposed to generate natural images by utilizing a stacked structure consisting of multiple generators and discriminators to generate images of different sizes. Tao \etal~\cite{Tao18attngan} employ the attention mechanism into this problem into their solution, which is able to synthesize images with fine-grained details from the text. Another line of literature concentrates on editing images by natural language description. For instance,~Dong \etal~\cite{dong2017semantic} manipulate images semantically with text descriptions. Nam~\etal~\cite{nam2018text} enhance fine-grained details by learning disentangled visual attributes from text-adaptive discriminator. However, most of them only succeed in the flower or bird image generation. In this paper, we present a text guided person image synthesis framework which can generate and edit the person pose and attribute according to natural language text while retaining the identity of the person.  

%
\vspace{-5pt}
\section{Method}
\vspace{-5pt}
\subsection{Problem Definition}
\vspace{-5pt}
Our goal is to simultaneously transfer the pose and the appearance of a person in the reference image corresponding to the given text description.

\noindent
\textbf{Training data. }
For each person in the training dataset, there is a tuple $(x, x', p, t)$ containing the source (reference) image $x$ and the target image $x'$ of the same identity with a different pose. $p$ and $t$ are the pose and the text description of $x'$, respectively.

\noindent
\textbf{Our pipeline. }
To tackle this challenging problem, we factorize it into two stages:
\vspace{-5pt}
\begin{itemize}
\item{\textbf{Stage-I:} We infer a reasonable pose based on the given text $t$. (See Sec.~\ref{sec:text_guided_pose})}
\vspace{-5pt}
\item{\textbf{Stage-II:} We generate a person image in which the pose and the attribute details of that person are changed according to target pose $p$ and text $t$. (See Sec.~\ref{sec:pose_and_attribute})}
\end{itemize}
\vspace{-5pt}
\subsection{Text Guided Pose Generator}
\label{sec:text_guided_pose}
\vspace{-5pt}

In Stage-I, we propose a novel approach (see Fig.~\ref{fig:model_1}), named \emph{text guided pose generator}, to infer a reasonable pedestrian pose satisfying the description.

We obtain the prior about poses in the training dataset as the basic poses and manipulate joints in these poses. Generally, the direction of a target pose is first estimated based on the description, then the target pose is generated in conjunction with detailed fine-tuning.

\noindent
\textbf{Basic poses. }
Synthesizing a pose directly from the text is difficult, as both the orientation and the other details (\eg, motions, posture) of a pose need to be considered. Following~\cite{qian2017pose}, we group the poses of all training images into $K$ clusters and compute the mean pose $p^\text{basic}_i$ of the $i$-th cluster, forming a basic pose set $\lbrace p^\text{basic}_i \rbrace_{i=1}^{K}$ (see Fig.~\ref{fig:model_1} for basic poses, where we use $K=8$ like~\cite{qian2017pose}). We assume that the basic poses orient toward all $K$ different directions.

\noindent
\textbf{Pose inference. }
Given the text description $t$ corresponding to the target image $x'$, we take the output of the final hidden layer of LSTMs as the sentence representation vector $\varphi_t$. We predict the orientation of the pose, $o = \mathop{\arg\max}_{o}F^{\text{ori}}(\varphi_t), o \in \lbrace 1,...,K \rbrace$. $F^{\text{ori}}$ is the orientation selection net implemented as fully-connected layers. The basic pose $p^\text{basic}_o$ which matches the orientation $o$ is selected from the $K$ basic poses.

We observe that verbs in the text can be vague in specifying the specific action of limbs. For example, the word \textit{walking} does not specify which leg to stride. The predicted pose by a regression method could either be striding on both legs or staying upright. Therefore, we train a single generator $G_\text{1}$ that learns to adjust the details of the pose, formulated as $G_\text{1}(p^\text{basic}_o, \varphi_t) \rightarrow \tilde{p}$. The discriminator $D_\text{1}$ outputs a probability that a pose is real conditioned on the text. $D_\text{1}$ forces $G_\text{1}$ to concern about posture details depicted by the text consistent with the real pose. The adversarial loss of discriminator $D_\text{1}$ is defined as
\begin{equation}\label{eq:pose_d}
\begin{aligned}
\mathcal{L}_{D_\text{1}} =& \ - \mathbb{E}_{p \sim \Pr (p)} \lbrack \log D_\text{1}(p) \rbrack \\
& \ - \mathbb{E}_{\tilde{p} \sim \Pr (\tilde{p})} \lbrack \log (1-D_\text{1}(\tilde{p})) \rbrack
\end{aligned}
\end{equation}
\begin{figure*}[tb]
\begin{center}
\includegraphics[width=\linewidth]{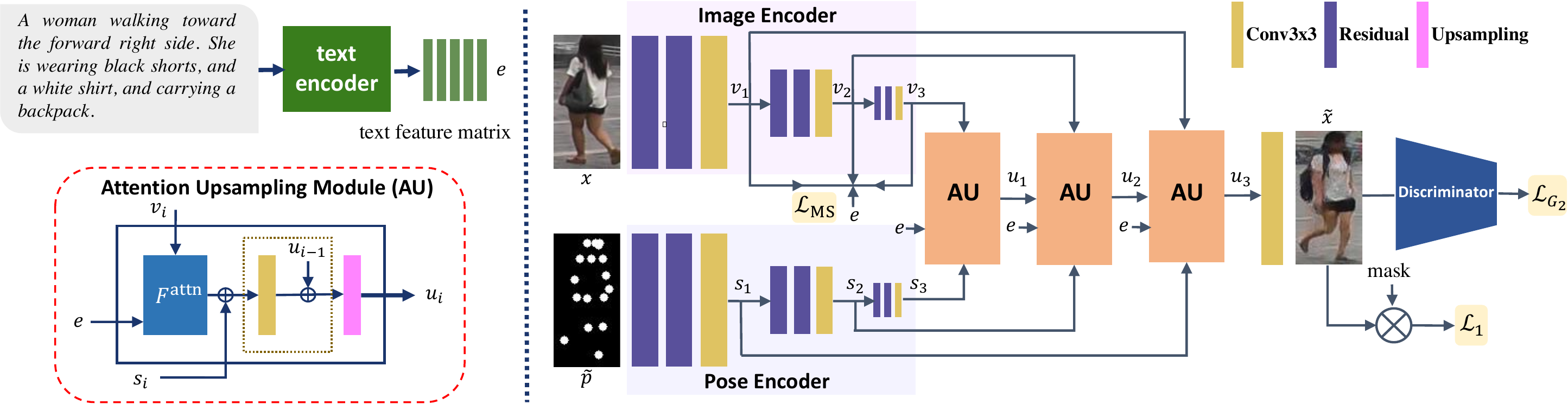}
\end{center}
\vspace{-13pt}
\caption{\textbf{Stage-II: Pose and attribute transferred person image generator. }It is a multi-modal learning scheme which builds the link between modalities of image, text, and pose. In addition, we propose a basic attentional upsampling (AU) module to better incorporate information of different modalities and spatial scales into image generation. The conjunction symbol in the AU module means the concatenation operation.}
\label{fig:model_2}
\end{figure*}

\noindent
And the adversarial loss of generator $G_\text{1}$ is 
\begin{equation}\label{eq:pose_g}
\mathcal{L}_{G_\text{1}}= - \mathbb{E}_{t\sim p_{data}} \lbrack \log D_\text{1}(G_\text{1}(p^\text{basic}_o, \varphi_t)) \rbrack
\end{equation}

However, we find that only using the adversarial loss makes the generated poses lack of pedestrian pose structure, as the values of pose heat-map are 1 merely within the radius while the rest are almost 0. Thus, we add a mean-square-error item, $\mathcal{L}_\text{mse}=\Vert \tilde{p} - p \Vert ^{2}$, to the adversarial loss of generator $G_\text{1}$, helping maintaining the unique structure.

The objective function of text guided pose generator is finally formulated as
\begin{equation}\label{eq:pose_total}
\mathcal{L}_\text{Stage-I}=\mathcal{L}_{G_\text{1}} + \lambda_{1} \mathcal{L}_\text{mse} + \lambda_{2} \mathcal{L}_\text{cls}
\end{equation}
\noindent
Here, $\lambda_{1}$ and $\lambda_{2}$ are hyper-parameters for balancing the three terms of Eq. \eqref{eq:pose_total}. $\mathcal{L}_\text{cls}$ is the cross entropy between the estimated orientation $o$ and the real orientation $o_\text{real}$. 
\vspace{-5pt}
\subsection{Pose and Attribute Transferred Person Image Generator}
\label{sec:pose_and_attribute}
\vspace{-5pt}

We have predicted the target pose $\tilde{p}$ based on the text $t$ so far. In Stage-II, our goal is to transfer the pose to the target pose $\tilde{p}$ and edit the appearance (\eg, cloth color) according to certain key words in description $t$. \footnote{The attribute that we focus on mainly is the color of clothes in this work, while in principle, our method can be easily extended to
accommodating other attributes.} To tackle this challenging problem, we propose a multi-task \emph{pose and attribute transferred image generator}, which is shown in Fig.~\ref{fig:model_2}.

Our multi-task person image synthesis framework is built upon the encoder-decoder structure.

\begin{itemize}
\vspace{-5pt}
\item{The \emph{image encoder} extracts the image feature maps $(v_1, v_2, ..., v_m)$ of different scales by taking the source image $x$ as input. $v_i \in \mathbb{R}^{l_i \times h_i \times w_i}$, where $l_i, h_i, w_i$ are the dimension, height, and width of the feature map at the $i$-th scale, $i \in [1,...m]$, $m$ is the total number of downsampling in the encoder.}
\vspace{-5pt}
\item{The \emph{text encoder} is a bi-directional LSTM which extracts the text feature matrix $e \in \mathbb{R}^{L \times N}$ of text $t$. $L$ is the dimension of hidden states, $N$ is the number of words. $e$ is concatenated by the hidden states $(h_1, h_2, ..., h_N)$ corresponding to every word in $t$.}
\vspace{-5pt}
\item{The \emph{pose encoder} \cite{ma2018disentangled} extracts the pose feature representations $(s_1, s_2, ..., s_m)$ of different scales by taking the target pose $\tilde{p}$ as input, similar to the image encoder, $s_i \in \mathbb{R}^{l_i \times h_i \times w_i}$.}
\vspace{-5pt}
\end{itemize}

\noindent
\textbf{Text-to-visual attention. }
We take the text feature matrix $e$ and the image feature map $v_i$ as input to calculate a dynamic text-to-visual attention which indicates the trend that each word takes care of each local visual region when generating images. The text-to-visual attention at the $i$-th scale is calculated as
\begin{equation}\label{f_attn}
F_{i}^{\text{attn}}(\hat{e}_i, \bar{v_i}) = \hat{e}_i \text{Softmax} (\hat{e}_i^\top \bar{v_i})
\end{equation}
\noindent
where the visual feature $v_i \in \mathbb{R}^{l_i \times h_i \times w_i}$ is reshaped to $ \bar{v}_i \in \mathbb{R}^{l_i \times h_i w_i}$, and the text feature matrix $e$ is converted to a common semantic space $\hat{e}_i$ by an embedding layer as $\hat{e}_i = W_i e, W_i \in \mathbb{R}^{l_i \times L}$.

\noindent
\textbf{Attentional upsampling. }
We propose a basic module, named attentional upsampling (AU). The motivation is that our pose and attribute transfer problem contains multiple modalities of data (image, pose, and text). We apply this module to better incorporate the text-to-visual attention features and the pose features at different scales. The pose features conduct the layout and the structure, while the text-to-visual attentional features integrate attribute information from words into visual representation. In our experiments, we observe that the module is capable of transferring the pose and the attribute appearance of a person in the source image, while keeping the invariant identity of the source image and the generated image.

Our attentional upsampling operates on the image feature maps and pose feature maps at the same scale (see Fig.~\ref{fig:model_2} Attentional Upsampling). To better retain information in the source image, the generators for image synthesizing and upsampling are weight-sharing, in which the fused features of different scales correspond to lower resolution to higher resolution, respectively. The total $m$ attentive operations in the upsampling correspond to those in the downsampling.

By using Eq. \eqref{f_attn}, we calculate the text-to-visual attention at the $i$-th scale as $z_i = F_{i}^{\text{attn}}(\hat{e}, \bar{v_i})$. Then, $z_i$, $s_i$ and the previous upsampling result $u_{i-1}$ are incorporated and upsampled by
\begin{equation}
u_i = F^{\text{up}}_{i}(z_i, s_i, u_{i-1})
\end{equation}

For the smallest scale (\ie, $i=1$),  $z_1$ and the pose feature $s_1$ are concatenated and upsampled as $u_1 = F^{\text{up}}_{1}(z_1, s_1)$. In such a recursive manner, the information of all different scales is incorporated in the final attentional upsampling result $u_m$. $u_m$ is passed through a ConvNet to output the generated image $\tilde{x}$. In practice, we implement $F^{\text{up}}$ as ConvNets with a nearest neighbor upsampling.

\noindent
\textbf{Multimodal loss. }
The multimodal loss function helps to establish the mapping between every word in the text and regions of images at different scales. The multimodal loss impose alignment among them for subsequently transferring appearance controlled by the text.

Similar to Eq. \eqref{f_attn}, the visual-to-text attention is calculated by
\begin{equation}\label{ci_attn}
c_i = \hat{v_i}  \text{Softmax} (\hat{v_i}^{\top} e)
\end{equation}

The visual feature $v_i$ is first reshaped to $\bar{v}_i \in \mathbb{R}^{l_i \times h_i w_i}$ and then converted to a common semantic space as $\hat{v}_i = U_i \bar{v}_i, U_i \in \mathbb{R}^{L \times l_i}$. $c_i \in \mathbb{R}^{L \times N}$, where the $j$-th column of $c_i$ denotes the visual text attention for the $j$-th word at the $i$-th scale.

Inspired by~\cite{Tao18attngan}, we calculate the similarity between the visual-to-text representation and the text feature matrix. The multi-scale visual-to-text distance is
\begin{equation}
\mathcal{D}(Q, T) = \sum_{i=1}^{m} \log \bigl ( \sum_{j=1}^{N} \text{exp} (r(c_{ij}, e_j)) \bigr )
\label{eq:vis_text_dis}
\end{equation}
\noindent
where $Q$ refers to the image (query) and $T$ refers to the description. $r(\cdot, \cdot)$ is the cosine similarity between two vectors, $m$ is the number of scales.

For a batch of our training pairs $\{ ( x_i, t_i ) \}_{i=1}^{I}$, we calculate the multi-scale visual-to-text distance matrix $\Lambda$; the element $\Lambda_{(i,j)}=\mathcal{D}(x_i, t_j)$. Following~\cite{Tao18attngan}, the posterior probability that the text $t_i$ matches with the image $x_i$ is calculated as $P(t_i \vert x_i) = \text{Softmax}(\Lambda)_{(i,i)}$. Similarly, the posterior that the image $x_i$ matches the text $t_i$ is $P(x_i \vert t_i) = \text{Softmax}(\Lambda ^ \top) _{(i,i)}$.

The multimodal similarity $\mathcal{L}_{\text{MS}}$ measures the interaction responses for the pairing of sentences and images in a batch
\begin{equation}
\mathcal{L}_{\text{MS}} = -\sum_{i=1}^{I} \log P(t_i \vert x_i) - \sum_{i=1}^{I} \log P(x_i \vert t_i)
\label{eq:ms_loss}
\end{equation}

\noindent
\textbf{Multi-task person image generator. }
The total objective function of our multi-task person image generator is defined as 
\begin{equation}
\mathcal{L}_\text{Stage-II}=\mathcal{L}_{G_\text{2}} + \gamma_{1} \mathcal{L}_1 + \gamma_{2} \mathcal{L}_{\text{MS}}
\label{eq:person_total}
\end{equation}
\noindent
where $\gamma_1$ and $\gamma_2$ are hyper-parameters. $\mathcal{L}_1$ is the L1 distance between generated image $\tilde{x}$ and real image $x'$, written as
\begin{equation}
\mathcal{L}_1 = \Vert (\tilde{x} - x') \odot M \Vert_{1}
\label{eq:person_l1}
\end{equation}
\noindent
where $M$ is the mask of the target pose~\cite{ma2017pose}. We use three conditional probabilities to improve the quality of the generated images. The adversarial loss for the generator $G_\text{2}$ is defined as
\begin{equation}\label{eq:person_g}
\footnotesize
\begin{aligned}
\mathcal{L}_{G_\text{2}} &= \underbrace{-\mathbb{E}_{\tilde{x} \sim \Pr(\tilde{x})} \lbrack \log D_\text{2}(\tilde{x}, e) \rbrack}_\text{text conditional loss} \;\underbrace{-\mathbb{E}_{\tilde{x} \sim \Pr(\tilde{x})} \lbrack \log D_\text{2}(\tilde{x}, p) \rbrack}_\text{pose conditional loss} + \\
&\;\;\;\; \underbrace{-\mathbb{E}_{\tilde{x} \sim \Pr(\tilde{x})} \lbrack \log D_\text{2}(\tilde{x}, e, p) \rbrack}_\text{text and pose conditional loss}
\end{aligned}
\end{equation}
\noindent
and the adversarial loss for the discriminator $D_\text{2}$ is
\begin{equation}\label{eq:person_d}
\footnotesize
\begin{aligned}
\mathcal{L}_{D_\text{2}} &= \underbrace{-\mathbb{E}_{x' \sim p_{data}} \lbrack \log D_\text{2}(x', e) \rbrack - \mathbb{E}_{\tilde{x} \sim \Pr(\tilde{x})} \lbrack \log (1-D_\text{2}(\tilde{x}, e) \rbrack}_\text{text conditional loss} + \\
&\;\;\;\; \underbrace{-\mathbb{E}_{x' \sim p_{data}} \lbrack \log D_\text{2}(x', p) \rbrack - \mathbb{E}_{\tilde{x} \sim \Pr(\tilde{x})} \lbrack \log (1-D_\text{2}(\tilde{x}, p) \rbrack}_\text{pose conditional loss} + \\
&\;\;\;\; \underbrace{-\mathbb{E}_{x' \sim p_{data}} \lbrack \log D_\text{2}(x', e, p) \rbrack - \mathbb{E}_{\tilde{x} \sim \Pr(\tilde{x})} \lbrack \log (1-D_\text{2}(\tilde{x}, e, p) \rbrack}_\text{text and pose conditional loss}
\end{aligned}
\end{equation}
\vspace{-5pt}
\subsection{VQA Perceptual Score}
\label{sec:vqa_perceptual_score}
\vspace{-5pt}

The evaluation metrics of GANs in the existing literature are not specifically designed for the attribute transfer task.
\begin{figure}[htp]
\begin{center}
\includegraphics[width=\linewidth]{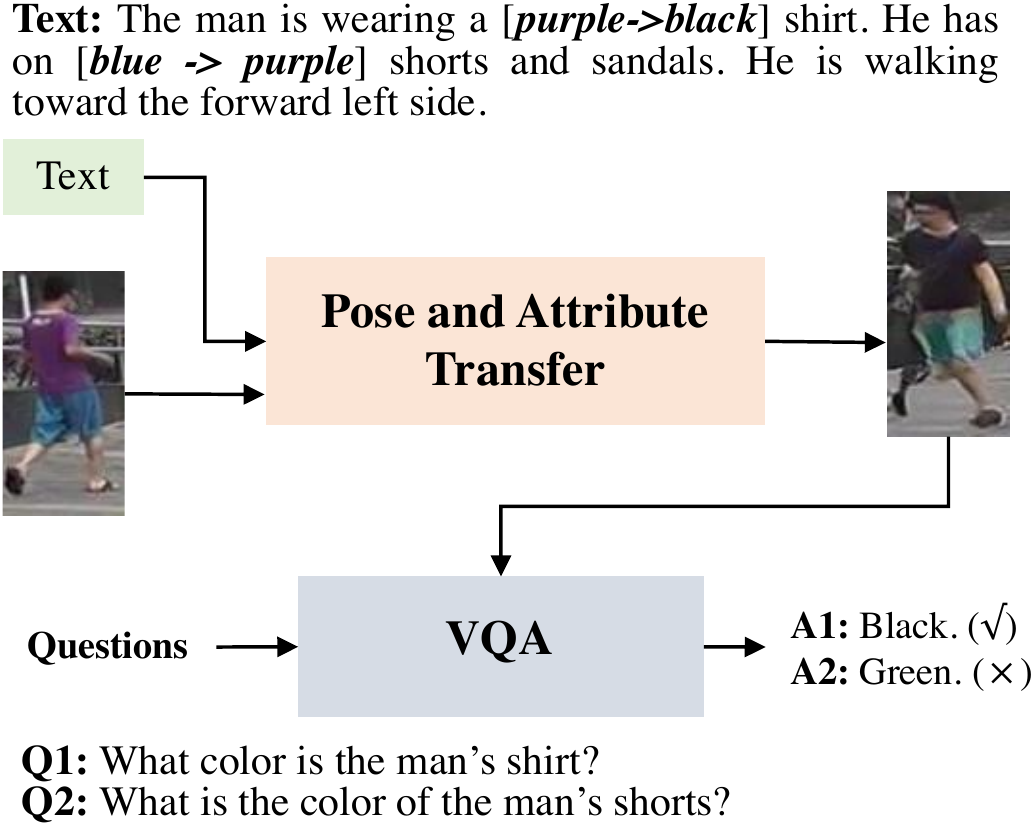}
\end{center}
\vspace{-8pt}
\caption{\textbf{Illustration of VQA perceptual Score.} The accuracy of answers returned by the VQA model denotes the attribute transfer correctness of generative models.}
\vspace{-10pt}
\label{fig:vqa_illustration}
\end{figure}

The Inception Score (IS)~\cite{salimans2016improved} measures the authenticity of synthesis and Structural Similarity (SSIM)~\cite{wang2004image} measures the structural integrity of images. To this end, we propose a novel metric, named VQA perceptual score, for the assessment of the attribute transfer correctness, \ie, whether the attributes of a person in the generated images are in agreement with the text description.

We first generate images using the method we propose by randomly changing the color adjectives of the clothes in the text (10 colors are considered). Correspondingly, the color word is recorded as the correct answer. Then a related question is automatically generated about the body part (shirt, pants, \etc) and its color. We ask the VQA model~\cite{kazemi2017show} with the question and the image. Finally, we gather the responses from the VQA model and calculate the accuracy, \ie, the VQA perceptual score. Assuming that $T$ is the number of images which receive all the correct answers from the VQA model, and that there are $N$ images in total, the VQA perceptual score is defined as $\frac{T}{N}$.

\section{Experiments}
\label{sec:exp}
\vspace{-5pt}

\subsection{Dataset}
\label{sec:exp_dataset}
\vspace{-5pt}

\noindent
\textbf{CUHK-PEDES dataset}~\cite{li2017person} is the only caption-annotated pedestrian image dataset as far as we know. The dataset contains 40,206 images of 13,003 persons collected from five person re-identification datasets, CUHK03~\cite{li2014deepreid}, Market-1501~\cite{zheng2015person}, SSM~\cite{xiao2016end}, VIPER~\cite{gray2007evaluating}, and CUHK01~\cite{li2012human}. Each image in the dataset is annotated with descriptions by crowd-sourcing.

In order to train the text guided pose generator, we add some phrases which describe the orientation, since the original descriptions rarely contain the information. An orientation phrase is an important guidance because otherwise the orientation in the generated image can be arbitrary and randomly different when lacking the orientation information. This may bring troubles to both the model training and testing.

For each image, a short phrase is appended according to the result of the clustering mentioned in Sec.~\ref{sec:text_guided_pose}. Every phrase corresponds to one of the $K=8$ basic orientations. We have manually checked the phrases to ensure a high quality dataset.

Following~\cite{ma2017pose}, the identities of training set and testing set are exclusive. All images in the dataset are resized to $128 \times 64$. In the training set, we have 149,049 pairs each of which is composed of two images of the same identity but different poses. We have 63,878 pairs in the testing set.

\vspace{-5pt}
\subsection{Comparison with Baselines}
\label{sec:exp_quati}
\vspace{-5pt}

As there is no existing work exactly comparable with this work, we implement four different baselines with appropriate modifications to make them comparable with our model as follows.\footnote{We do not use any extra pre-trained models in our framework, such that all the parameters in our model are trained from scratch, which is different from~\cite{Tao18attngan}.}

\begin{enumerate}
\vspace{-5pt}
\item
{Modified Semantic Image Synthesis (SIS)~\cite{dong2017semantic} (mSIS). }SIS uses a plain text encoder without our proposed attentional upsampling module. SIS only edits the attribute, but the pose is not involved. We append a pose encoder to it for pose transfer. The generator synthesizes a new person image based on the encoded reference image feature and the conditional representations of the target pose and the target text description. 
\vspace{-5pt}
\item
{Modified AttnGAN~\cite{Tao18attngan} (mAttnGAN). }We add an image encoder and a pose encoder to the original AttnGAN~\cite{Tao18attngan, nam2018text}. Specifically, an additional {\tt inception\_v3} network is adopted to establish the link among different words in the text and the sub-regions in the image. 
\vspace{-5pt}
\item
{Modified $\text{PG}^\text{2}$~\cite{ma2017pose} (m$\text{PG}^\text{2}$). }Pose guided person image generation ($\text{PG}^\text{2}$) only generates the pose transferred person image. In this baseline, we append a text encoder for attribute transfer. Our multi-task problem is separated into two single-task problems, in which the pose transferred image is first synthesized and then the image is edited according to the text description step by step.
\vspace{-5pt}
\item
{Single attentional upsampling (SAU). }It conducts only $m=1$ attentional upsampling module at the smallest scale, serving as an ablation study for our complete attentional upsampling modules. 
\vspace{-5pt}
\end{enumerate}

\noindent
\textbf{Quantitative analysis. }
Following~\cite{ma2018disentangled}, we use the Inception Score (IS)~\cite{salimans2016improved} and the Structural Similarity (SSIM)~\cite{wang2004image} to measure the quality of generated person images. We evaluate the IS on tasks of pose transfer (PT) and pose and attribute transfer (P\&AT). We only evaluate the SSIM on PT, as the SSIM is calculated based on images' mean and variance values which vary during attribute transferring.

We evaluate the four baselines and our method on metrics of IS and SSIM, as shown in Table~\ref{tab:ssim_and_is}. We can see that mSIS, mAttnGAN, and m$\text{PG}^\text{2}$ are the improved variants of the existing methods while their IS and SSIM values are lower than that of our model. It indicates that simply replenishing the rest procedure of the existing methods may not be feasible for the challenging problem proposed in this paper. SAU is better than the other baselines, while it is also worse than our complete framework. It indicates that the attentional upsampling module proposed in this work enables a robust learning of pose and attribute transferring.

\begin{figure*}[t]
\begin{center}
\includegraphics[width=\linewidth]{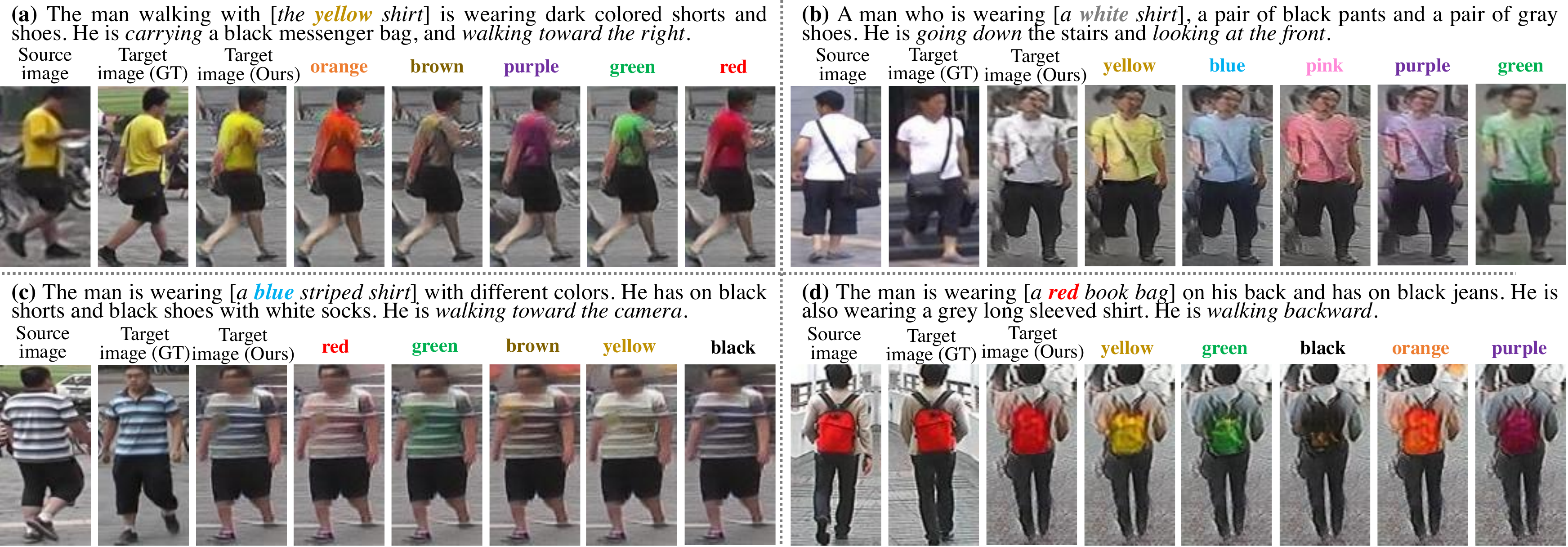}
\end{center}
\vspace{-10pt}
\caption{Four examples of text guided person image synthesis by our model. The first two columns are reference and target images (\ie, ground truth (GT)) from the training set. The third column is the target image generated by our model. We additionaly show that our model is capable of transferring attribute if we dedicate to changing the attribute (\eg, color) in the description.}
\label{fig:colorful_trans}
\end{figure*}
\begin{figure*}[t]
\begin{center}
\includegraphics[width=\linewidth]{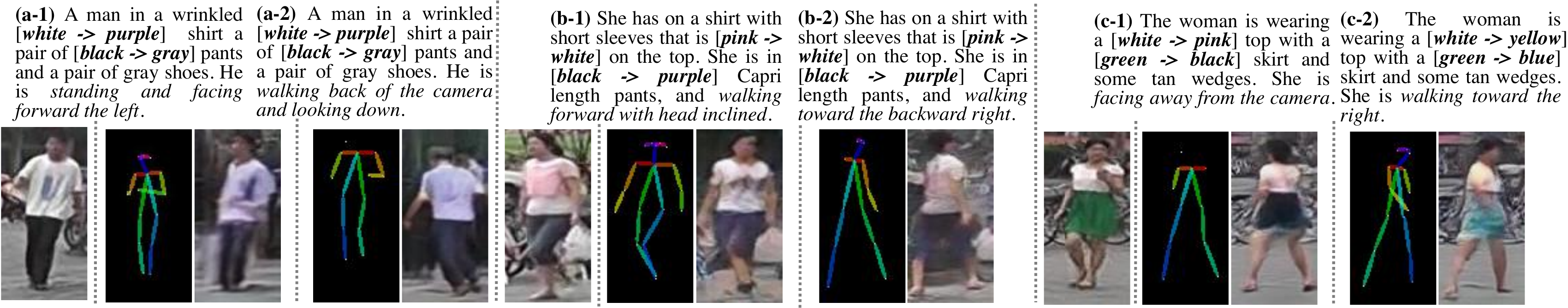}
\end{center}
\vspace{-10pt}
\caption{Interactive editing. By arbitrarily changing input words, our model can change a person to different poses for the same reference image. Our model can also transfer different attributes in one image at the same time, indicating a flexible and robust image generation procedure.}
\label{fig:interactive_3}
\end{figure*}
\begin{table}[h]
\vspace{-5pt}
\small
\begin{center}
\begin{tabular}{l c c c c}
\toprule 
\footnotesize{\bf{Model}} & \footnotesize{\bf{SSIM~(PT)}} & \footnotesize{\bf{IS~(PT)}} & \footnotesize{\bf{IS~(P\&AT)}}\\
\midrule
\footnotesize{mSIS} & \footnotesize{0.239}~\scriptsize{$\pm$~.106} & \footnotesize{3.707}~\scriptsize{$\pm$~.185} & \footnotesize{3.790}~\scriptsize{$\pm$~.182}\\
\footnotesize{mAttnGAN} & \footnotesize{0.298}~\scriptsize{$\pm$~.126} & \footnotesize{3.695}~\scriptsize{$\pm$~.110} & \footnotesize{3.726}~\scriptsize{$\pm$~.123}\\
\footnotesize{m$\text{PG}^\text{2}$} & \footnotesize{0.273}~\scriptsize{$\pm$~.120} & \footnotesize{3.473}~\scriptsize{$\pm$~.009} & \footnotesize{3.486}~\scriptsize{$\pm$~.125}\\
\footnotesize{SAU} & \footnotesize{0.305}~\scriptsize{$\pm$~.121} & \footnotesize{4.015}~\scriptsize{$\pm$~.009} & \footnotesize{4.071}~\scriptsize{$\pm$~.149}\\
\midrule
\footnotesize{Ours} & \footnotesize{\textbf{0.364}}~\scriptsize{$\pm$~.123} & \footnotesize{\textbf{4.209}}~\scriptsize{$\pm$~.165} & \footnotesize{\textbf{4.218}}~\scriptsize{$\pm$~.195}\\
\bottomrule
\end{tabular}
\end{center}
\vspace{-15pt}
\caption{The SSIM score for pose transfer, and the IS for pose transfer and pose \& attribute transfer (higher is better).}
\vspace{-5pt}
\label{tab:ssim_and_is}
\end{table}

\noindent
\textbf{VQA perceptual score. }
The VQA perceptual score of our model and the baselines are shown in Table~\ref{tab:vqa_perceptual_score}. mSIS gains a relatively high score, while its generated images almost lose the human structure which is intolerable for visual effects. The scores of mAttnGAN and m$\text{PG}^\text{2}$ are relatively low, confirming that a separate training of the two tasks is hard to achieve a balance between transfers of pose and attribute. Our model jointly addresses the two tasks with a multi-scale module to achieve competititve results.

\begin{table}[!h]
\begin{center}
\vspace{-5pt}
\footnotesize
\begin{tabular}{l c c}
\toprule 
\textbf{Model} & \textbf{VQA score} \\
\midrule
mSIS & 0.275 \\
mAttnGAN & 0.139 \\
m$\text{PG}^\text{2}$ & 0.110 \\
SAU & 0.205 \\
\midrule
Ours & \textbf{0.334} \\
\bottomrule
\end{tabular}
\end{center}
\vspace{-15pt}
\caption{VQA perceptual score (higher is better).}
\vspace{-8pt}
\label{tab:vqa_perceptual_score}
\end{table}

\vspace{-5pt}
\subsection{Qualitative Results}
\label{sec:exp_quali}
\vspace{-5pt}

Fig.~\ref{fig:colorful_trans} shows that our Stage-II framework generates realistic person images based on the text description and the predicted poses from Stage-I. In fact, the text guided image synthesis can be regarded as a semi-supervised problem, as there are only image-text pairs in the training dataset without ground-truth images corresponding to different text descriptions w.r.t the same identity. Nevertheless, by editing the descriptive words of various pedestrian appearance parts (\eg, shirt, pants), our model is able to accurately change these parts in the image generation procedure. It indicates that our model is able to capture sufficient information from the text, while holding an effective control on the text.

Our two-stage based method can edit both the pose and the attribute of the identity in the reference image by the natural language description, which is an interactive editing process for users. Fig.~\ref{fig:interactive_3} shows the results of the predicted poses and the generated images. Our method enhances the ability of the text for both the pose and the attribute interpolation. We can even change more than one word about the color attribute in the text description, and the synthesis is reasonable and correct corresponding to the text.

\begin{figure}[tp]
\begin{center}
\includegraphics[width=\linewidth]{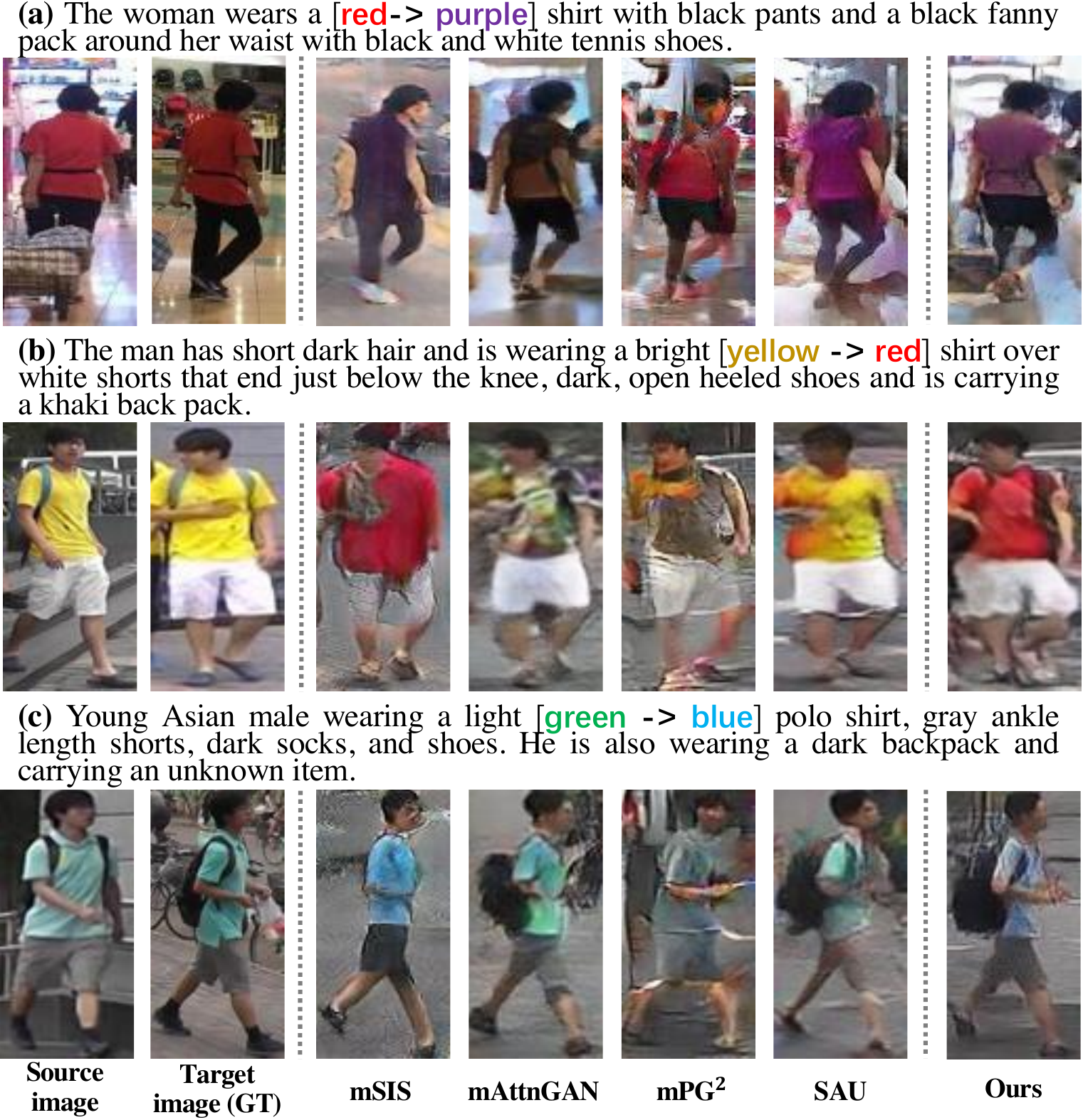}
\end{center}
\vspace{-8pt}
\caption{Qualitative comparison of our method and the baselines. Our method generates more valid and vivid person images.}
\vspace{-10pt}
\label{fig:baseline_compare}
\end{figure}

Fig.~\ref{fig:baseline_compare} shows a qualitative comparison of our method and different baselines. In the experiment, we find that there is a trade-off among identity invariance, pose transfer, and attribute transfer. For instance, m$\text{PG}^\text{2}$ first changes the pose, and then transfers the attribute of the person. However, the better the pose changes, the more difficult m$\text{PG}^\text{2}$ is to transfer the attribute. It is mainly because the distribution of the optimal pose is different from the distribution of the optimal attribute. This is also pointed out by~\cite{donahue2016adversarial} that the distinction of learned distribution may hurt the learning of feature representations when generating images. It is worth mentioning that although SAU adopts a single attentional module, its results are relatively better than those of the other baselines. However, SAU only integrates embedded features of different modalities at the smallest scale. In the experiment, we observe that this leads to more unreal attribute transfer. Thus, we use $m=3$ attentional upsampling in our complete framework. Intuitively, the multi-scale upsampling module exploits receptive fields of different ranges to learn visual-word mapping on diverse spatial scales so as to better generate more realistic details. (\eg, the woman's pack is preserved by our method in Fig.~\ref{fig:baseline_compare}(a).)
\begin{figure}[tp]
\centering
\includegraphics[width=\linewidth]{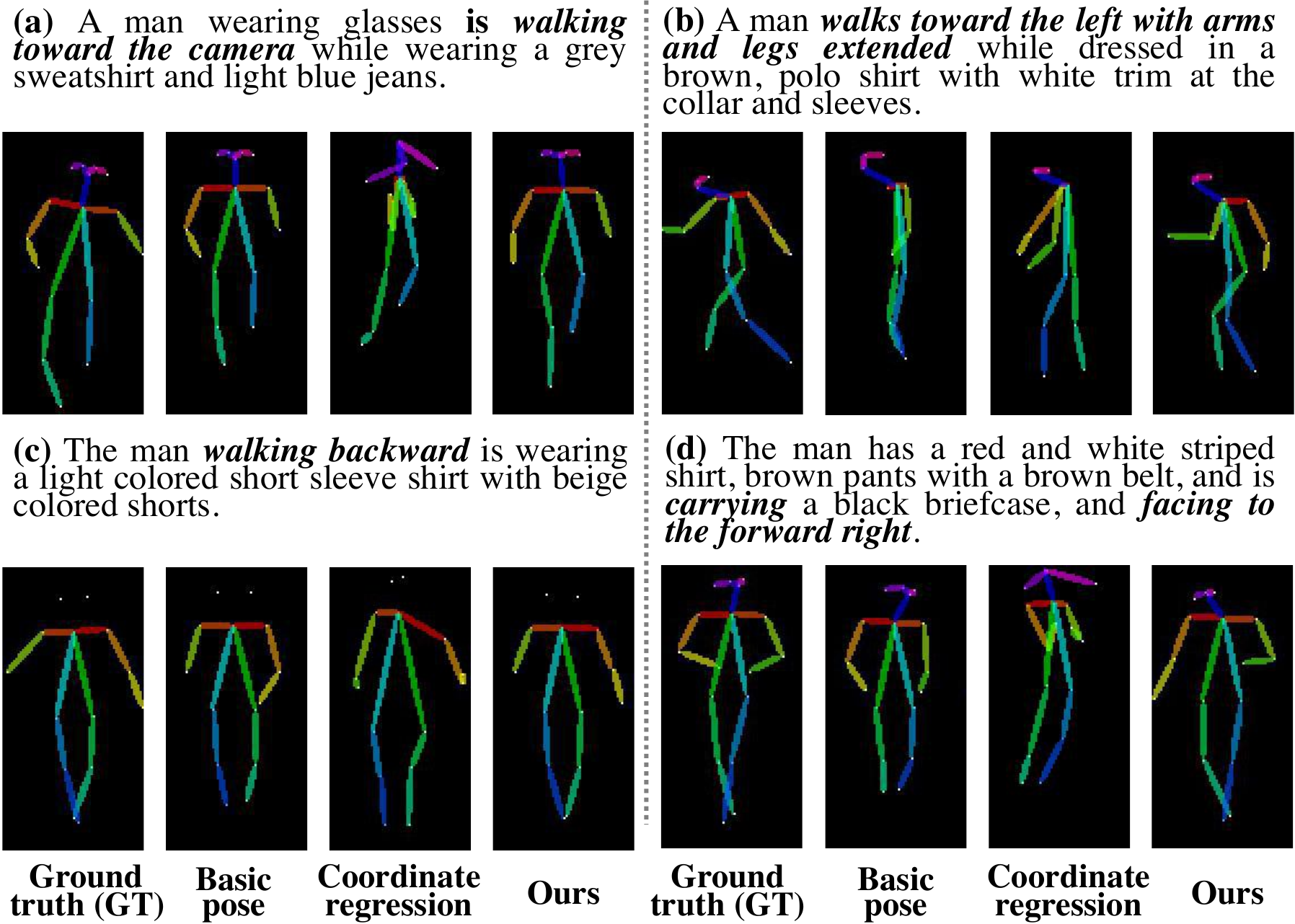}
\vspace{-15pt}
\caption{Qualitative comparison between our text guided pose generator and the coordinate regression method. The coordinate regression method may result in some distortions of joints, and our text guided pose generator generates more reasonable poses.}
\vspace{-15pt}
\label{fig:pose_fig}
\end{figure}

\noindent
\textbf{Pose inference using GANs. }
Fig.~\ref{fig:pose_fig} shows the selected basic poses and the inferred poses given text descriptions. The inferred poses are largely different from the basic poses, and our Stage-I model is able to concentrate on specific key words in the text (\eg, walking, carrying) as these key words imply large changes in the posture of specific body parts (\eg, arms, legs). Our model learns to adjust these details so that the inferred poses are much closer to the real ones, providing precise target poses for subsequent procedure of person image generation. We also implement a coordinate regression method as the baseline.  As shown in Fig.~\ref{fig:pose_fig}, the coordinate regression method may lead to the distortion of some joints. 
\vspace{-5pt}
\section{Conclusion}
\vspace{-5pt}

In this paper, we present a novel two-stage pipeline to manipulate the visual appearance (pose and attribute) of a person image based on natural language descriptions. The pipeline first learns to infer a reasonable target human pose based on the description, and then synthesizes an appearance transferred person image according to the text in conjunction with the target pose. Extensive experiments show that our method can interactively exert control over the process of person image generation by natural language descriptions. 

\noindent
\textbf{Acknowledgments. }
This work is supported in part by NSFC (61672456), Zhejiang Lab (2018EC0ZX01-2), the fundamental research funds for central universities in China (No. 2017FZA5007), Artificial Intelligence Research Foundation of Baidu Inc., the Key Program of Zhejiang Province, China (No. 2015C01027), the funding from HIKVision, and ZJU Converging Media Computing Lab.
{\small
\bibliographystyle{ieee_fullname}
\bibliography{egbib}
}


\end{document}